\documentclass[runningheads]{llncs}
\usepackage[T1]{fontenc}
\usepackage{graphicx}
\usepackage{multirow}%
\usepackage[hidelinks]{hyperref}
\usepackage[inkscapelatex=false]{svg}
\begin{document}
\title{BiLSTM-VHP: BiLSTM-Powered Network for Viral Host Prediction}
\author{Azher Ahmed Efat\inst{1, 2} \and
Farzana Islam\inst{3} \and
Annajiat Alim Rasel\inst{2} \and
Munima Haque\inst{3}}
\authorrunning{A.A. Efat et al.}
%
\institute{Department of Computer Science, Iowa State University, Ames, IA, 50010, USA \email{efat@iastate.edu} \and
Department of Computer Science and Engineering, Brac University, Dhaka, Bangladesh\\
 \email{azher.ahmed.efat@g.bracu.ac.bd}, \email{annajiat@gmail.com} \and
Biotechnology Program, Department of Mathematics and Natural Sciences, Brac University, Dhaka, Bangladesh\\
\email{\{islam.farzana,munima.haque\}@bracu.ac.bd}}
%
%

%
\maketitle              
\begin{center}
\textbf{Note:} This preprint has not undergone peer review or any post-submission improvements or corrections. 
The peer-reviewed Version of Record of this contribution is published in \textit{International Conference on Advances in Distributed Computing and Machine Learning}, 
and is available online at \href{https://doi.org/10.1007/978-981-96-6721-5_10}{https://doi.org/10.1007/978-981-96-6721-5\_10}
\end{center}
\begin{abstract}
Recorded history shows the long coexistence of humans and animals, suggesting it began much earlier. Despite some beneficial interdependence, many animals carry viral diseases that can spread to humans. These diseases are known as zoonotic diseases. Recent outbreaks of SARS-CoV-2, Monkeypox and swine flu viruses have shown how these viruses can disrupt human life and cause death. Fast and accurate predictions of the host from which the virus spreads can help prevent these diseases from spreading. This work presents BiLSTM-VHP, a lightweight bidirectional long short-term memory (LSTM)-based architecture that can predict the host from the nucleotide sequence of orthohantavirus, rabies lyssavirus, and rotavirus A with high accuracy. The proposed model works with nucleotide sequences of 400 bases in length and achieved a prediction accuracy of 89.62\% for orthohantavirus, 96.58\% for rotavirus A, and 77.22\% for rabies lyssavirus outperforming previous studies. Moreover, performance of the model is assessed using the confusion matrix, F-1 score, precision, recall, microaverage AUC. In addition, we introduce three curated datasets of orthohantavirus, rotavirus A, and rabies lyssavirus containing 8,575, 95,197, and 22,052 nucleotide sequences divided into 9, 12, and 29 host classes, respectively. The codes and dataset are available at \url{https://doi.org/10.17605/OSF.IO/ANFKR}

\end{abstract}

\section{Introduction}\label{sec1}

Animals play various roles in human society and are an integral part of human history. From being pets to being livestock, those may stay in direct contact with humans in different ways. Alien animals have been identified as zoonotic hosts of one or more zoonoses in studies \cite{b2} and they can transmit those viruses to humans once they come into contact. With deforestation, humans and domestic animals are more likely to be exposed to wild animals. Humans are also in risk of infectious diseases including coming into the contact of rabies lyssavirus, ebola virus, hantavirus due to their adventerous nature \cite{b26}. Due to their evolvability, transmissibility, and lack of therapeutic options, zoonotic viruses pose a great threat \cite{b3} and can occasionally cause pandemics and havoc. Furthermore, emerging diseases are likely to be associated with zoonotic pathogens\cite{b4}, and some changes in human environments may favor these pathogens. In the past, an enormous number of people died due to outbreaks of coronavirus, swine flu, and other viruses. To address an outbreak, it is crucial to find the origin host and isolate it as soon as possible.\\ 
Continuous research is ongoing in predicting the hosts of viruses.  In recent studies, various machine learning algorithms have been utilized to identify influenza A virus origin hosts \cite{b6}.  Various representations of the genome has been used to predict the host taxonomic information \cite{b8} \cite{b9}. Additionally, a random forest model can identify the host of coronaviruses based on the whole genome sequence and spike protein sequence \cite{b10}. Other studies have shown the use of SVM and the Mahalanobis distance to predict the host of coronaviruses \cite{b11}. In addition, the hosts of rabies and influenza A viruses were also predicted using SVM \cite{b12}. ML models have also been used to predict GBS host \cite{b28}.
\\
However, traditional machine learning algorithms suffer from imbalanced host classes and a lack of annotated virus information. To anticipate the hosts of newly discovered viruses, data scarcity and class imbalance were addressed using transfer learning and ensemble learning \cite{b13}. Additionally, deep neural networks and ensemble classifier were used to identify the host species for viruses \cite{b14} \cite{b35} \cite{b34}. In addition, studies such as VIDHOP provide DNN-based architectures that can predict viral hosts for specific viruses and can also be trained for new viruses \cite{b16}. Although much work has been done on predicting the hosts of novel viruses including SARS-CoV-2 and other common zoonotic viruses such as influenza A virus, very little work has focused on other zoonotic viruses such as orthohantavirus, rabies lyssavirus, and the rotavirus A. HFRS caused by hantavirus is appearing as a serious threat to public health \cite{b30}. Additionally, countless children loss their lives worldwide because of rotavirus each year \cite{b31}.\\
Viral host prediction task has its own significances for many reasons. By knowing which animals are likely to be hosts, health authorities can monitor these species for signs of new or emerging infections. Early detection of viruses in animal hosts can act as a warning for possible human outbreaks, enabling preemptive actions to prevent spillover events. Furthermore, understanding the range of hosts a virus can helps scientists anticipate mutations and design vaccines that offer broader protection. The contributions of this research are as follows:
\begin{itemize}
     \item The introduction of three curated datasets containing nucleotide sequences for rotavirus A, rabies lyssavirus, and orthohantavirus, along with their respective hosts.
     \item A lightweight bidirectional long short-term memory (LSTM)-based model to predict the origin host of rotavirus A, orthohantavirus, and rabies lyssavirus.
    \item To the best of our knowledge, this is the first work on host prediction for orthohantavirus from nucleotide sequences.
    \item Our proposed models outperform those of previous studies in predicting the origin host from nucleotide sequences for rotavirus A as well as for rabies lyssavirus and trained on more nucleotide sequences including latest sequences compared to those of previous studies.
\end{itemize}

\section{Methodology}\label{sec2}

The prediction of the viral host from nucleotide sequences is split into three phases: (1) dataset curation; (2) data preprocessing; and (3) model creation and training.

\subsection{Dataset curation}\label{subsec1}
For this research, we curated three datasets containing nucleotide sequences with host details for orthohantavirus (taxid:1980442), rabies lyssavirus (taxid:11292), and rotavirus A (taxid:28875) from the NCBI virus database \cite{b17}. While curating the dataset, the hosts with 100+ nucleotide sequences were retained. Table \ref{Table 1} displays an overview of these three datasets. The term “Host” refers to the host organism from which the virus was isolated, as provided by the sequence submitter \cite{b17}. 

\begin{table}[b]
\caption{Overview of Orthohantavirus, Rotavirus A, and Rabies Lyssavirus dataset}\label{Table 1}%
\centering
\begin{tabular}{|l|l|l|}
\hline
{Dataset} & {\# of Hosts} & {\# of Nucleotide Sequences} \\
\hline
Orthohantavirus & \multicolumn{1}{r|}{9}   & \multicolumn{1}{r|}{8,575} \\ 
Rabies Lyssavirus    & \multicolumn{1}{r|}{29} & \multicolumn{1}{r|}{22,052}  \\ 
Rotavirus A    & \multicolumn{1}{r|}{12}& \multicolumn{1}{r|}{95,197}  \\ 
\hline
\end{tabular}
\end{table}

\subsection{Dataset Preprocessing}\label{subsec2}
Data preprocessing is a crucial part of any deep learning task to reduce future potential errors. The preprocessing task is divided into the following phases: (1) resizing the sequence length; (2) handling and encoding sequence characters.
\subsubsection{Resizing the sequence length}\label{subsubsec1}

Nucleotide sequences for viruses vary in length. To train the model, we needed to standardize the sequence length. We trained the model on lengths of 400, 600, and 1000 bases. The 1000-base length was selected as it encompassed more than half of the sequences across all datasets. However, no significant improvement was observed with longer sequences. Thus, we chose 400 bases, as it performed competitively while reducing computational costs and providing sufficient length to extract feature. All sequences were resized to 400 bases: sequences longer than 400 bases were truncated, and shorter ones were padded by repeating the original sequence until it reached 400 bases.
\subsubsection{Handling and Encoding Sequence Characters}\label{subsubsec2}
There are four types of nucleotide bases in a nucleotide sequence: ``A'', ``G'', ``C'', and ``T''. Moreover, any unknown characters in a nucleotide sequence are treated as ``N''. All the characters that were not ``A'', ``G'', ``C'', ``T'' or ``N'' were replaced with ``N''. The nucleotide bases of the sequences were encoded using one hot sequence. The encoding was performed as: ``A'' = [1,0,0,0,0], ``C'' = [0,1,0,0,0], ``G'' = [0,0,1,0,0], ``T'' = [0,0,0,1,0], and ``N'' = [0,0,0,0,1].

\subsection{Model creation and training}\label{subsec3}
\subsubsection*{Model creation}\label{subsubsec4}
While working with nucleotide sequences, previous information or bases are significant. A RNN is a kind of neural network that can retain information from the past and work with sequential data. However, RNNs suffer from long-term dependency in sequence and the vanishing gradient problem. LSTM addresses these problems faced by RNNs. Furthermore, bidirectional LSTM allows processing of the input sequence from both directions. This research proposes a bidirectional LSTM-based neural network model for predicting the viral hosts of orthohantavirus, rotavirus A, and rabies lyssavirus. The proposed BiLSTM-VHP model consists of an input layer, one BiLSTM layer, a dropout layer, a batch normalization layer, and two dense layers. The model's output layer, which is the final dense layer, has a variable unit size dependent on the number of classes. The dropout layer has been used to prevent overfitting as the datasets are imbalanced with the dropout rate 0.2. The architecture of BiLSTM-VHP for each of the viral host prediction tasks is shown in Table \ref{Table 2}. 

\begin{table}[t]
\caption{Architecture of BiLSTM-VHP for the viral host prediction task\label{Table 2}}
\begin{tabular}{@{\extracolsep{\fill}}|l|c|c|c|@{\extracolsep{\fill}}}
\hline%
& \multicolumn{3}{@{}c|@{}}{Output Shape} \\
\cline{2-4}%
Layer & Orthohantavirus & Rabies Lyssavirus & Rotavirus A \\
\hline
input\_layer (InputLayer)& [(None, 400, 5)]& [(None, 400, 5)] & [(None, 400, 5)] \\
BiLSTM\_Model (Bidirectional) & (None, 256) & (None, 256) & (None, 256) \\
DROPOUT (Dropout) & (None, 256) & (None, 256) & (None, 256) \\
BatchNormalized (BatchNormalization)  & (None, 256) & (None, 256) & (None, 256)     \\
Dense\_Layer1 (Dense) & (None, 64) & (None, 64) & (None, 64) \\
output (Dense)  &  (None, 9) &  (None, 29) &  (None, 12) \\
\hline
Total parameters: & 155,273 & 156,573 & 155,468\\
Trainable parameters:& 154,761 & 156,061 & 154,956\\
Non-trainable parameters:& 512 & 512 &512\\
\hline
\end{tabular}
\end{table}

\subsubsection{Model training}\label{subsubsec5}
\begin{figure}[t]
\includegraphics[width=\textwidth]{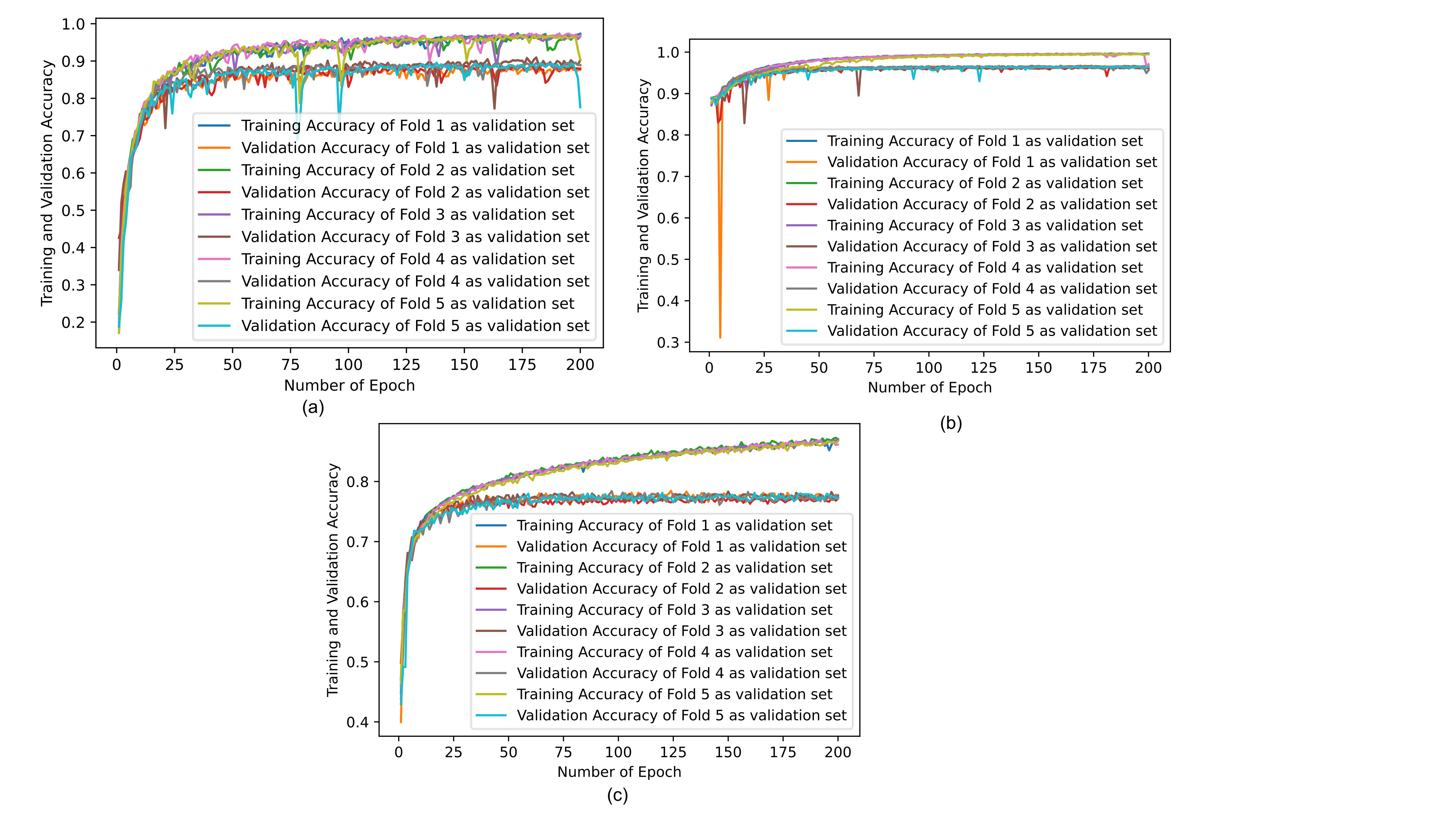}
 \caption{{5-fold training vs. validation accuracy graph for (a)Orthohantavirus, (b)Rotavirus A, (c)Rabies Lyssavirus host prediction.}
  }
\label{Fig 1}
\end{figure}

The three datasets were initially divided into 80\% training and 20\% test datasets prior to the model being trained. The input data that the model uses determine how well it performs. Therefore, it is crucial to evaluate the model's efficiency on unobserved data. Five fold cross-validation was applied on training data to assess the performance of the models during training. The model was trained for 200 epochs for each training-validation combination, and the weight of the model that had the best validation accuracy was saved. The training was stopped after 200 epochs, as there was no significant improvement in the validation accuracy. The training vs. validation accuracy graph for each of the viral host prediction tasks is displayed in Fig \ref{Fig 1}. Moreover, hyperparameter tuning is a critical step in the development of deep learning models. The hyperparameters used in model training are given in Table \ref{Table 3}. In addition to this, one of the part of training a deep learning architecture is to address the class imbalance. To address class imbalance, we trained our model in two setting: (1) trained the model without any special treatment, (2) adding class weight. Class weight assigns different weight to the classes that influence the loss function during training and gives higher importance to the minority class. The equation to calculate the class weights is given in equation \ref{eq1}. Here $w_{j}$ is the weight of the class j and $\#samples_{j}$ is the number of samples in the class j. After training the model on two of the settings, we found that the model performs better with class weight for orthohantavirus. However, the model performed better without any additional treatment for rotavirus A and rabies lyssavirus. Moreover, because of extreme class imbalance, we kept sampling out of study. 

\begin{equation}
w_{j} =  \frac{\#samples}{\#classes * \#samples_{j}}\label{eq1}
\end{equation}
\begin{table}[t]
\caption{{Hyperparameters of BiLSTM-VHP for the viral host prediction} \label{Table 3}}%
\centering
\begin{tabular}{@{\extracolsep{\fill}}|l|l|@{\extracolsep{\fill}}}
\hline
{Parameter Name} & {Value} \\ \hline
batch size & 128 \\

loss & sparse\_categorical\_crossentropy\\ 
metrics  & accuracy    \\ 
optimizer& adam\\
Dropout rate & 0.2\\\hline
\end{tabular}
\end{table}

\section{Results}\label{sec3}

This section discusses the results achieved from evaluating and validating the performance of the proposed model on the three curated datasets. The evaluation factors are the confusion matrix, f-1 score, recall, precision, microaverage
AUC and accuracy. Assessing the model's performance on unobserved data is crucial for understanding how it might handle sequences it wasn’t trained on. The similarity score between the training and test datasets serves as a metric for evaluating the model's effectiveness on new data. Thus, a similarity comparison was performed between the training and test datasets before and after preprocessing using global pairwise sequence alignment, k-mer frequency analysis with k=4, and sequence identity. The similarity score in Table \ref{Table 4} shows that the training and test datasets are highly dissimilar to each other. Additionally, a chi-squared test was performed to compare the distribution of nucleotides in the training and test data by setting the significance level at 0.05. Table \ref{Table 5} shows the significant difference between the training and test datasets.

\begin{table}[t]

\caption{
{Similarity comparison of the train and test dataset} \label{Table 4}}

\begin{tabular}{|l|l|l|l|l|}
\hline%
Comparison step
                              & Dataset & \begin{tabular}[c]{@{}l@{}}Sequence  alignment\end{tabular}  & \begin{tabular}[c]{@{}l@{}}K-mer Frequency\\ Analysis\end{tabular} &\begin{tabular}[c]{@{}l@{}}Sequence \\ identity\end{tabular}   \\ \hline
\multirow{3}{*}{\begin{tabular}[c]{@{}l@{}}Before \\ Preprocessing\end{tabular} } & orthohantavirus   & \multicolumn{1}{r|}{-43.04\%}                    & \multicolumn{1}{r|}{41.16\%}& \multicolumn{1}{r|}{20.86\%}                   \\  \cline{2-5}
                                      & rotavirus A       &\multicolumn{1}{r|}{-6.24\%}                   & \multicolumn{1}{r|}{40.26\%}                          & \multicolumn{1}{r|}{23.38\%}                    \\ \cline{2-5} 
                                      & rabies lyssavirus & \multicolumn{1}{r|}{-45.59\%}                    & \multicolumn{1}{r|}{40.18\%}                          & \multicolumn{1}{r|}{22.62\%}                    \\ \hline
\multirow{3}{*}{\begin{tabular}[c]{@{}l@{}}After \\ Preprocessing\end{tabular} }  & orthohantavirus   & \multicolumn{1}{r|}{48.69\%}                  & \multicolumn{1}{r|}{39.99\%}                  & \multicolumn{1}{r|}{27.52\%}                  \\  \cline{2-5}
                                      & rotavirus A       & \multicolumn{1}{r|}{54.58\%}                     & \multicolumn{1}{r|}{40.00\%}                           & \multicolumn{1}{r|}{29.27\%}                    \\\cline{2-5}  
                                      & rabies lyssavirus & \multicolumn{1}{r|}{58.89\%}                     & \multicolumn{1}{r|}{40.00\%}                           & \multicolumn{1}{r|}{30.29\%}                    \\ \hline
\end{tabular}
\end{table}

\begin{figure}[t]%
\centering
\includegraphics[width=\textwidth]{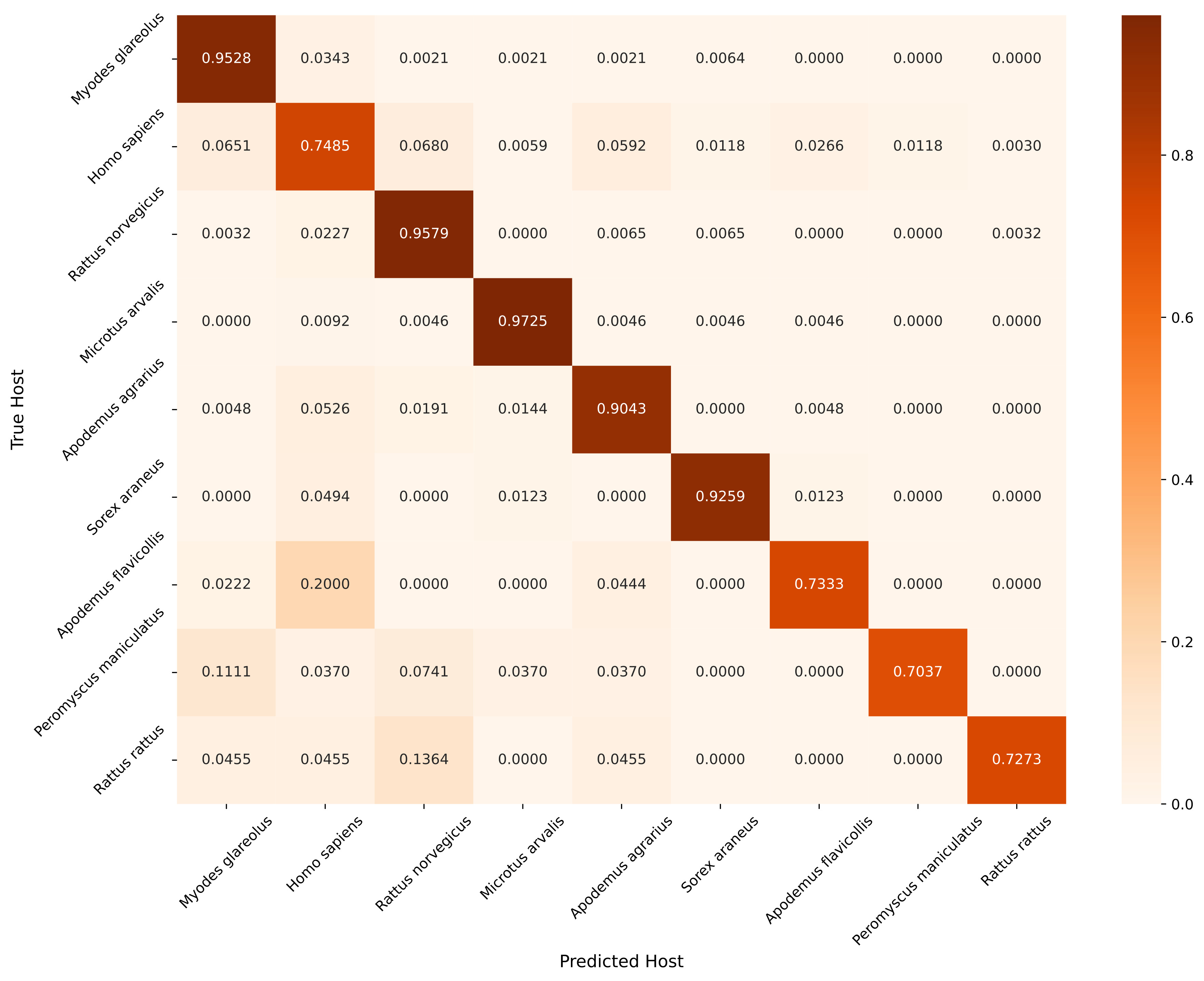}
\caption{{Confusion matrix of BiLSTM-VHP for orthohantavirus host prediction}}\label{Fig 2}
\end{figure}

\begin{figure}[]%
\centering
\includegraphics[width=\textwidth]{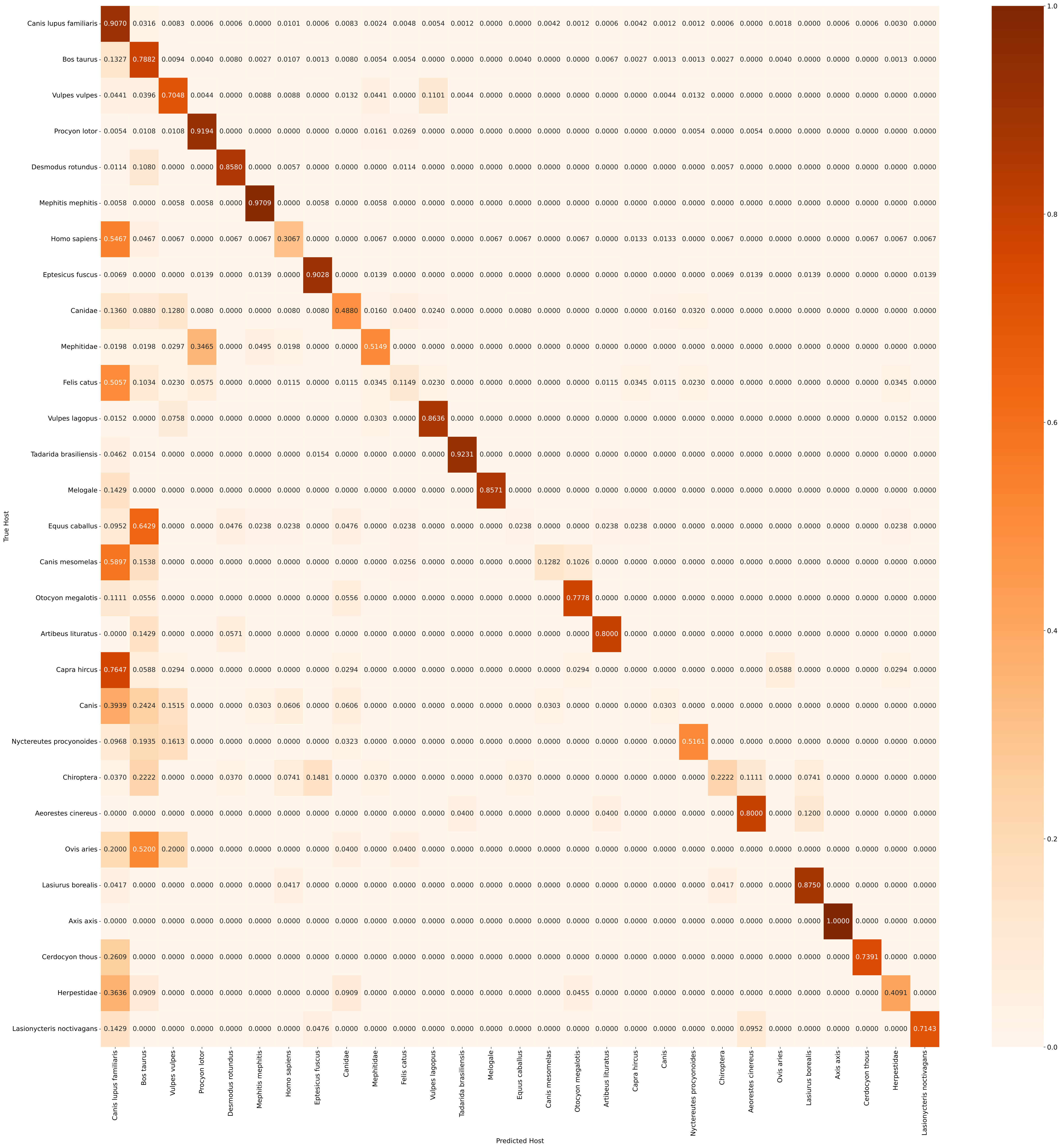}
\caption{{ Confusion matrix of BiLSTM-VHP for rabies lyssavirus host prediction}}\label{Fig 3}
\end{figure}

\begin{figure}[t]%
\centering
\includegraphics[width=0.88\textwidth]{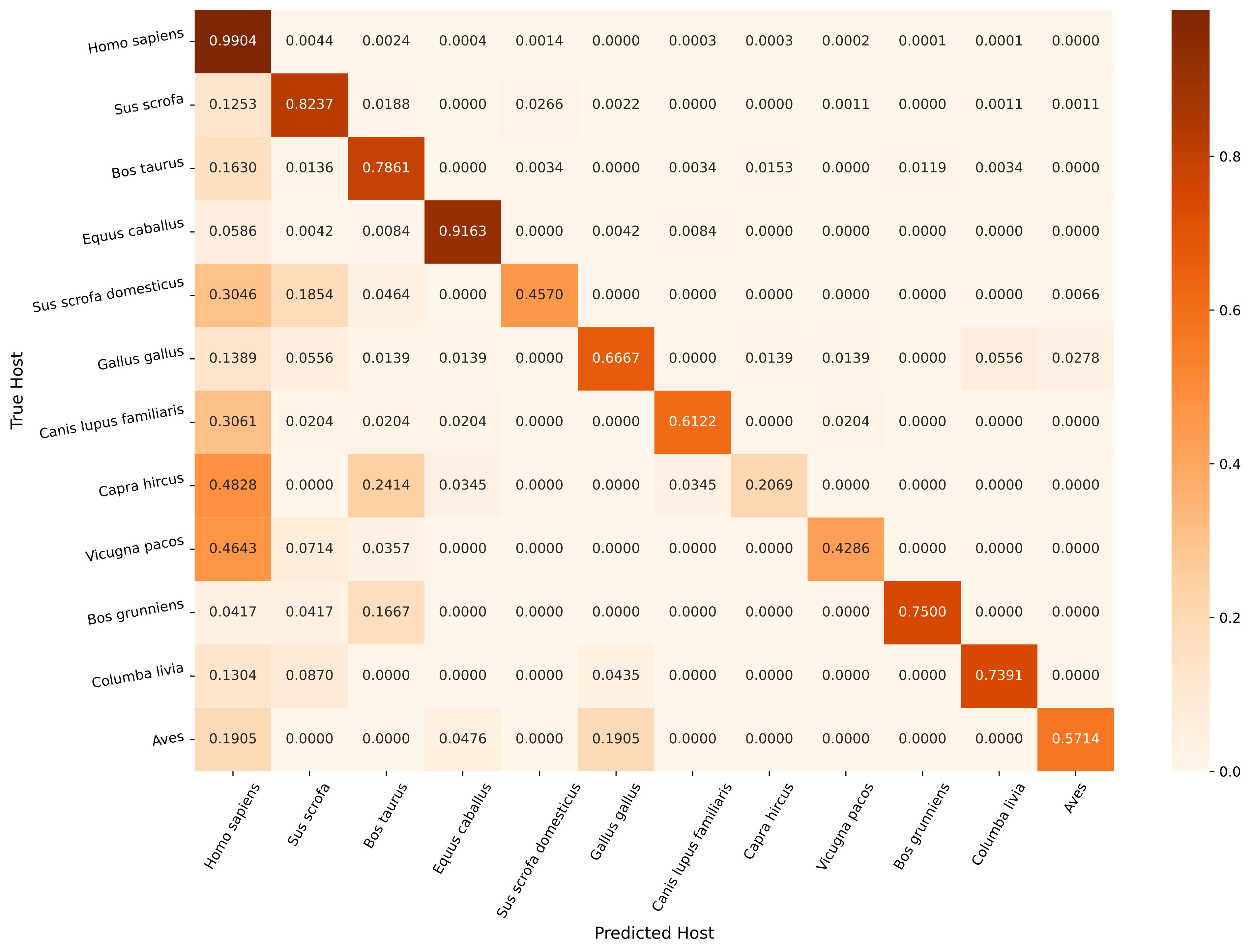}
\caption{{ Confusion matrix of BiLSTM-VHP for rotavirus A host prediction}}\label{Fig 4}
\end{figure}

\begin{table}[t]

\caption{
{Comparison of the train and test dataset using chi-squared test} \label{Table 5}}
\begin{tabular}{@{\extracolsep{\fill}}|l|l|l|l|l|@{\extracolsep{\fill}}}
\hline%
{Comparison step}                             & {Dataset}  & {Chi-squared statistic} & \begin{tabular}[c]{@{}l@{}}Degrees of \\ freedom\end{tabular}& {p-value \hspace{1em} } \\ \hline
\multirow{3}{*}{Before Preprocessing} & orthohantavirus   & \multicolumn{1}{r|}{197.81}                  & \multicolumn{1}{r|}{4}    & $1.11\times 10^{-41} $                    \\ 
                                      & rotavirus A       & \multicolumn{1}{r|}{190.99}                     & \multicolumn{1}{r|}{4}    & $3.25\times 10^{-40} $                      \\ 
                                      & rabies lyssavirus & \multicolumn{1}{r|}{78.03}                   & \multicolumn{1}{r|}{4}  & $4.56\times 10^{-16} $                    \\ \hline
\multirow{3}{*}{After Preprocessing}  & orthohantavirus   & \multicolumn{1}{r|}{287.95}                  & \multicolumn{1}{r|}{4}    & $4.31\times 10^{-61} $                   \\ 
                                      & rotavirus A       & \multicolumn{1}{r|}{47.62}                     & \multicolumn{1}{r|}{4}                     & $1.13\times 10^{-09} $                     \\  
                                      & rabies lyssavirus & \multicolumn{1}{r|}{88.36}                    & \multicolumn{1}{r|}{4}                 & $2.93\times 10^{-18} $                      \\ \hline
\end{tabular}
\end{table}
After testing the model on the curated datasets, we found that the proposed model achieved an accuracy of 89.62\% for orthohantavirus, 96.58\% for rotavirus A, and 77.22\% for rabies lyssavirus. Additionally, the confusion matrices of BiLSTM-VHP for the host prediction tasks is shown in Figs \ref{Fig 2}–\ref{Fig 4}. The confusion matrices show that the proposed model was capable of learning all the classes of orthohantavirus and rotavirus A. For Orthohantavirus, it had over 90\% accuracy on 5 of the classes and more than 70\% accuracy for rest of the classes each. Similarly, we can see that the model had over 70\% accuracy for 6 of the classes of rotavirus A including 99.04\% accuracy for Homo sapiens. However, the model struggled to learn the class \textit{Capra hircus} for rotavirus A. Moreover, the confusion matrix for rabies lyssavirus shows that the model struggled slightly for some classes even though it had over 70\% accuracy for 17 classes. This is due to the high number of classes, as the model is trained on 29 classes for the rabies lyssavirus. Furthermore, the precision, recall, and F-1 scores for the three viral host prediction tasks are given in Table \ref{Table 6}. From Table \ref{Table 6}, we see that the model has high F1 accuracy for rotavirus A and orthohantavirus. The model has low recall and F1 score for few classes of rabies lyssavirus due to data scarcity. Additionally, BiLSTM-VHP has achieved microaverage AUCs of 0.9888, 0.9980, and 0.9818 on the Orthohantavirus, Rotavirus A, and Rabies Lyssavirus datasets respectively, showing that the model's overall performance is reliable.

\begin{table}[]
\caption{{Precision, recall, and f-1 score of BiLSTM-VHP  }\label{Table 6}}%
\begin{tabular}{@{}|l|l|l|l|l|@{}}
\hline
{Virus}&{Host class}&{Precision} & {Recall} & {F-1 score} \\ \hline
  \multirow{3}{*}{Rotavirus A }    &    \textit{Aves} &        \multicolumn{1}{r|}{0.75}  &        \multicolumn{1}{r|}{0.57}   &       \multicolumn{1}{r|}{0.65}        \\\cline{2-5} 
     &    \textit{Bos grunniens} &       \multicolumn{1}{r|}{0.69}  &        \multicolumn{1}{r|}{0.75}   &       \multicolumn{1}{r|}{0.72}        \\ \cline{2-5}
    &     \textit{Bos taurus} &    \multicolumn{1}{r|}{0.85} &         \multicolumn{1}{r|}{0.79}   &    \multicolumn{1}{r|}{0.82}       \\ \cline{2-5}
& \textit{Canis lupus familiaris} &        \multicolumn{1}{r|}{0.75}   &       \multicolumn{1}{r|}{0.61}    &      \multicolumn{1}{r|}{0.67}        \\ \cline{2-5}
& \textit{Capra hircus}  &         \multicolumn{1}{r|}{0.29}   &       \multicolumn{1}{r|}{0.21}    &      \multicolumn{1}{r|}{0.24}        \\ \cline{2-5}
 & \textit{Columba livia} &        \multicolumn{1}{r|}{0.65}  &        \multicolumn{1}{r|}{0.74}   &       \multicolumn{1}{r|}{0.69}        \\ \cline{2-5}
&  \textit{Equus caballus} & \multicolumn{1}{r|}{0.96}  &        \multicolumn{1}{r|}{0.92}    &      \multicolumn{1}{r|}{0.94}       \\ \cline{2-5}
&  \textit{Gallus gallus}   &       \multicolumn{1}{r|}{0.86}   &       \multicolumn{1}{r|}{0.67}   &       \multicolumn{1}{r|}{0.75}        \\ \cline{2-5}
 
	& \textit{Homo sapiens}     &  \multicolumn{1}{r|}{0.98}    &  \multicolumn{1}{r|}{0.99}  &        \multicolumn{1}{r|}{0.99}     \\ \cline{2-5}
    &      \textit{Sus scrofa} &   \multicolumn{1}{r|}{0.86}   &   \multicolumn{1}{r|}{0.82}   &       \multicolumn{1}{r|}{0.84}       \\ \cline{2-5}

   &        \textit{Sus scrofa domesticus} &       \multicolumn{1}{r|}{0.58}  &        \multicolumn{1}{r|}{0.46}   &       \multicolumn{1}{r|}{0.51}       \\ \cline{2-5}

& \textit{Vicugna pacos} &       \multicolumn{1}{r|}{0.67} &         \multicolumn{1}{r|}{0.43}  &        \multicolumn{1}{r|}{0.52}        \\\hline

         \multirow{9}{*}{Orthohantavirus } &
\textit{Apodemus agrarius}      &    \multicolumn{1}{r|}{0.87}  &    \multicolumn{1}{r|}{0.90}  &    \multicolumn{1}{r|}{0.89}   \\   \cline{2-5}
       &    \textit{Apodemus flavicollis}  &    \multicolumn{1}{r|}{0.73}  &    \multicolumn{1}{r|}{0.73}  &    \multicolumn{1}{r|}{0.73}    \\  \cline{2-5}
 &\textit{Homo sapiens}   &    \multicolumn{1}{r|}{0.83}   &   \multicolumn{1}{r|}{0.75}  &    \multicolumn{1}{r|}{0.79}    \\ \cline{2-5}
       &   \textit{Microtus arvalis}            &     \multicolumn{1}{r|}{0.96}   &   \multicolumn{1}{r|}{0.97}  &    \multicolumn{1}{r|}{0.97}   \\  \cline{2-5}   
	   & \textit{Myodes glareolus}      &    \multicolumn{1}{r|}{0.94}   &   \multicolumn{1}{r|}{0.95}    &  \multicolumn{1}{r|}{0.95}   \\   \cline{2-5}
    &     \textit{Peromyscus maniculatus}    &   \multicolumn{1}{r|}{0.83}  &    \multicolumn{1}{r|}{0.70}  &    \multicolumn{1}{r|}{0.76}  \\   \cline{2-5}    
 &\textit{Rattus rattus}     &  \multicolumn{1}{r|}{0.89}  &    \multicolumn{1}{r|}{0.73}  &    \multicolumn{1}{r|}{0.80}     \\   \cline{2-5}
 &\textit{Rattus norvegicus}    &     \multicolumn{1}{r|}{0.90}  &    \multicolumn{1}{r|}{0.96}  &    \multicolumn{1}{r|}{0.93}      \\ \cline{2-5}

        &    \textit{Sorex araneus}             &    \multicolumn{1}{r|}{0.88}   &   \multicolumn{1}{r|}{0.93}  &    \multicolumn{1}{r|}{0.90}  \\  \hline

 \multirow{29}{*}{Rabies Lyssavirus }   &    \textit{Aeorestes cinereus}   &    \multicolumn{1}{r|}{0.71}  &    \multicolumn{1}{r|}{0.80}   &   \multicolumn{1}{r|}{0.75}        \\ \cline{2-5}
&\textit{Artibeus lituratus}    &   \multicolumn{1}{r|}{0.76}     & \multicolumn{1}{r|}{0.80}   &   \multicolumn{1}{r|}{0.78}        \\ \cline{2-5}
&\textit{Axis axis}   &    \multicolumn{1}{r|}{0.96}    &  \multicolumn{1}{r|}{1.00} &     \multicolumn{1}{r|}{0.98}        \\ \cline{2-5}
&\textit{Bos taurus}     &  \multicolumn{1}{r|}{0.76}   &   \multicolumn{1}{r|}{0.79}      & \multicolumn{1}{r|}{0.77}       \\ \cline{2-5}
&\textit{Canis}  & \multicolumn{1}{r|}{0.10}   &   \multicolumn{1}{r|}{0.03}   &   \multicolumn{1}{r|}{0.05}        \\ \cline{2-5}
&\textit{Canidae}  &     \multicolumn{1}{r|}{0.64}   &   \multicolumn{1}{r|}{0.49}  &    \multicolumn{1}{r|}{0.55}       \\ \cline{2-5}
	&  \textit{Canis lupus familiaris}     &  \multicolumn{1}{r|}{0.81}   &   \multicolumn{1}{r|}{0.91}   &   \multicolumn{1}{r|}{0.85}      \\ \cline{2-5}
&\textit{Canis mesomelas}  &     \multicolumn{1}{r|}{0.38}    &  \multicolumn{1}{r|}{0.13} &     \multicolumn{1}{r|}{0.19}        \\ \cline{2-5}
&\textit{Capra hircus}      & \multicolumn{1}{r|}{0.00}   &   \multicolumn{1}{r|}{0.00} &     \multicolumn{1}{r|}{0.00}        \\ \cline{2-5}
&\textit{Cerdocyon thous}    &   \multicolumn{1}{r|}{0.89}   &   \multicolumn{1}{r|}{0.74} &      \multicolumn{1}{r|}{0.81}        \\ \cline{2-5}
& \textit{Chiroptera}     &  \multicolumn{1}{r|}{0.46}    &  \multicolumn{1}{r|}{0.22}  &    \multicolumn{1}{r|}{0.30}        \\ \cline{2-5}
& \textit{Desmodus rotundus}     &  \multicolumn{1}{r|}{0.92}   &   \multicolumn{1}{r|}{0.86}   &   \multicolumn{1}{r|}{0.89}       \\ \cline{2-5}
&\textit{Eptesicus fuscus}   &    \multicolumn{1}{r|}{0.93}   &   \multicolumn{1}{r|}{0.90}     & \multicolumn{1}{r|}{0.92}       \\ \cline{2-5}
&\textit{Equus caballus}  &     \multicolumn{1}{r|}{0.14}    &  \multicolumn{1}{r|}{0.02} &     \multicolumn{1}{r|}{0.04}        \\ \cline{2-5}
 &\textit{Felis catus}    &   \multicolumn{1}{r|}{0.27}  &    \multicolumn{1}{r|}{0.11}  &    \multicolumn{1}{r|}{0.16}        \\ \cline{2-5}
& \textit{Herpestidae} &      \multicolumn{1}{r|}{0.41} &     \multicolumn{1}{r|}{0.41}  &   \multicolumn{1}{r|}{0.41}        \\ \cline{2-5}
&  \textit{Homo sapiens}     &  \multicolumn{1}{r|}{0.55}    &  \multicolumn{1}{r|}{0.31}  &    \multicolumn{1}{r|}{0.39}       \\ \cline{2-5}
&  \textit{Lasionycteris noctivagans } &      \multicolumn{1}{r|}{0.83}    &  \multicolumn{1}{r|}{0.71}    &  \multicolumn{1}{r|}{0.77}        \\  \cline{2-5}
 & \textit{Lasiurus borealis}  &     \multicolumn{1}{r|}{0.75}    &  \multicolumn{1}{r|}{0.88}    &  \multicolumn{1}{r|}{0.81}        \\ \cline{2-5}
   
&\textit{Melogale}  &     \multicolumn{1}{r|}{0.98}   &   \multicolumn{1}{r|}{0.86}  &    \multicolumn{1}{r|}{0.91}        \\ \cline{2-5}
&\textit{Mephitidae}     &  \multicolumn{1}{r|}{0.61}   &   \multicolumn{1}{r|}{0.51}   &   \multicolumn{1}{r|}{0.56}       \\ \cline{2-5}
&   \textit{Mephitis mephitis} & \multicolumn{1}{r|}{0.92} & \multicolumn{1}{r|}{0.97} &     \multicolumn{1}{r|}{0.95}       \\ \cline{2-5}
 &   \textit{Nyctereutes procyonoides\ } &      \multicolumn{1}{r|}{0.55}  &    \multicolumn{1}{r|}{0.52} &     \multicolumn{1}{r|}{0.53}        \\ \cline{2-5}
 &   \textit{Otocyon megalotis} &       \multicolumn{1}{r|}{0.76}  &    \multicolumn{1}{r|}{0.78} &     \multicolumn{1}{r|}{0.77}        \\ \cline{2-5}
 &   \textit{Ovis aries}  & \multicolumn{1}{r|}{0.00}  &    \multicolumn{1}{r|}{0.00}  &    \multicolumn{1}{r|}{0.00}        \\ \cline{2-5}
&    \textit{Procyon lotor}      & \multicolumn{1}{r|}{0.78}    &  \multicolumn{1}{r|}{0.92}  &    \multicolumn{1}{r|}{0.84}       \\ \cline{2-5}
 &    \textit{Tadarida brasiliensis} &      \multicolumn{1}{r|}{0.94}  &    \multicolumn{1}{r|}{0.92}    &  \multicolumn{1}{r|}{0.93}        \\ \cline{2-5}
 &     \textit{Vulpes lagopus}   &    \multicolumn{1}{r|}{0.59}    &  \multicolumn{1}{r|}{0.86}   &   \multicolumn{1}{r|}{0.70}        \\ \cline{2-5}
   &        \textit{Vulpes vulpes}      & \multicolumn{1}{r|}{0.70}   &   \multicolumn{1}{r|}{0.70}  &    \multicolumn{1}{r|}{0.70}       \\ \hline
             
\end{tabular}
\end{table}


\section{Discussions}\label{sec4}
Along with assessing the proposed model on the three curated datasets, we compared it with models from previous studies and validated its results using an external dataset. To the best of our knowledge, no prior work has predicted the viral host of Orthohantavirus from nucleotide sequences. However, F.Mock et al. proposed two deep learning-based architectures- based on bidirectional LSTM, and CNN+bidirectional LSTM- that can be trained for viral host classification \cite{b16}. We trained both of the architectures proposed by F.Mock et al. with the default configuration on our curated orthohantavirus dataset for 200 epochs with early stops enabled and compared the results using accuracy. The comparison of performance is given in Table \ref{Table 9}. Moreover, both of the architectures of VIDHOP have numerous hidden layers, which might have contributed to the poor results, as the complexity of the model increases with increasing layers. In contrast, BiLSTM-VHP is very lightweight with 1,000,000 less parameters than the VIDHOP model and is not prone to overfitting.
\begin{table}[t]
\caption{{BiLSTM-VHP vs VIDHOP on orthohantavirus host prediction}\label{Table 9}}%
\centering
\begin{tabular}{@{\extracolsep{\fill}}|l|l|@{\extracolsep{\fill}}}
\hline
{Model} &{Accuracy }   \\ \hline
BiLSTM-VHP&  \multicolumn{1}{r|}{89.62\%}    \\ 
VIDHOP Bidirectional LSTM& \multicolumn{1}{r|}{32.80\%}     \\ 
VIDHOP CNN+Bidirectional LSTM& \multicolumn{1}{r|}{83.60\%}   \\ \hline
\end{tabular}
\end{table}

Additionally, there have been three similar works on the viral host classification of the rabies lyssavirus. S.Ali et al. classified rabies lyssavirus from nucleotide sequences into 12 host classes \cite{b23}. The authors curated 20,051 nucleotide sequences from RABV-GLUE, which contain nucleotide sequences from NCBI. On the other hand, our dataset has 22,052 nucleotide sequences divided into 29 host classes, curated from the NCBI virus database. Moreover, Z. Ming et al. \cite{b33} and F.Mock et al. \cite{b16} proposed a model for rabies lyssavirus host prediction which were trained on only 17 classes using the dataset proposed in \cite{b16}. Whereas our proposed BiLSTM-VHP is trained on 29 classes. Moreover, F.Mock et al. proposed another trained model and dataset to predict the host of rotavirus A, which is trained on 6 classes compared to our proposed BiLSTM-VHP, which is trained on 12 classes \cite{b16}. As much as we are aware of, the work of F.Mock et al. is the only previous work on the classification of viral hosts from nucleotide sequences for rotavirus A. We compared our model's performance with that of VIDHOP \cite{b16} and validated our model with their dataset. Thus, to compare the models, we tested both of the models on a subset of our test dataset and a subset of VIDHOP's test dataset containing the classes that are common to both of these studies. The comparative performance shown in Table \ref{Table 11}. We found that the proposed BiLSTM-VHP model outperformed previous models in every task even after being trained on more classes. 
\begin{table}[]
\caption{{BiLSTM-VHP vs VIDHOP on Rotavirus A and Rabies Lyssavirus }\label{Table 11}}%
\centering
\begin{tabular}{|l|l|l|l|}
\hline
{Virus}&{Dataset}& {BiLSTM-VHP} & {VIDHOP}    \\ \hline
\multirow{2}{*}{Rotavirus A} & \begin{tabular}[c]{@{}l@{}}Accuracy on the subset \\ of our test dataset\end{tabular} &  \multicolumn{1}{r|}{0.97}  &    \multicolumn{1}{r|}{0.96}   \\ \cline{2-4}
&\begin{tabular}[c]{@{}l@{}}Accuracy on the subset \\ of VIDHOP test dataset\end{tabular}& \multicolumn{1}{r|}{0.88}   &   \multicolumn{1}{r|}{0.86}  \\ \hline

\multirow{2}{*}{\begin{tabular}[c]{@{}l@{}}Rabies\\ Lyssavirus\end{tabular} }&\begin{tabular}[c]{@{}l@{}}Accuracy on the subset \\ of our test dataset\end{tabular}   &  \multicolumn{1}{r|}{0.74}  &   \multicolumn{1}{r|}{0.63}   \\ \cline{2-4}
&\begin{tabular}[c]{@{}l@{}}Accuracy on the subset \\ of VIDHOP test dataset\end{tabular}     & \multicolumn{1}{r|}{0.73}   &   \multicolumn{1}{r|}{0.73}  \\ \hline
\end{tabular}
\end{table}

\section{Conclusion}\label{sec13}

Zoonotic viruses pose a great threat, and if not taken care of promptly, they can occasionally cause pandemics. In this research, we presented a lightweight BiLSTM-based deep learning network, that can identify the viral hosts of rabies lyssavirus, orthohantavirus, and rotavirus A with high accuracy. BiLSTM-VHP achieved a prediction accuracy of 89.62\% for orthohantavirus, 96.58\% for rotavirus A, and 77.22\% for rabies lyssavirus outperforming the models used in previous studies. In addition, we introduced three curated datasets of orthohantavirus, rotavirus A, and rabies lyssavirus containing nucleotide sequences and host classes. In the future, balancing viral host-sequence data could be a crucial step for advancing this field. Additionally, future research will examine the geographical distribution of virus hosts and the timing of data collection to gain deeper insights into the virus's evolution, and changes in infection trends.

\begin{credits}
\subsubsection{\discintname}
 The authors have no competing interests to declare. 
\end{credits}

%

\end{document}